\PassOptionsToPackage{table,xcdraw}{xcolor}
\documentclass[acmsmall,screen]{acmart}
\pdfoutput=1
\setcopyright{none}

\usepackage{geometry}
\usepackage{hyperref}
\usepackage[nameinlink]{cleveref}
\usepackage{caption}
\usepackage{subcaption}
\usepackage{soul}
\usepackage{tablefootnote}

\graphicspath{ {./figures/} }

\acmJournal{PACMHCI}
\acmVolume{9}
\acmNumber{CSCW}
\acmArticle{}
\acmYear{2025}
\acmMonth{03}
\copyrightyear{}
\acmPrice{}
\acmDOI{}
\acmISBN{}

\makeatletter
\let\@authorsaddresses\@empty
\makeatother

\begin{document}

\title[]{The AI Double Standard: Humans Judge All AIs for the Actions of One}

\author{Aikaterina Manoli}
\orcid{0000-0003-2562-0380}
\affiliation{%
  \institution{Sentience Institute}
  \country{US}
}
\affiliation{%
  \institution{Max Planck institute for Human Cognitive and Brain Sciences}
  \country{Germany}
}
\email{katerina@sentienceinstitute.org}

\author{Janet V. T. Pauketat}
\orcid{0000-0003-3280-3345}
\affiliation{%
  \institution{Sentience Institute}
  \country{US}
}
\email{janet@sentienceinstitute.org}

\author{Jacy Reese Anthis}
\orcid{0000-0002-4684-348X}
\affiliation{%
  \institution{Sentience Institute}
  \country{US}
}
\affiliation{%
  \institution{Stanford University}
  \country{US}
}
\affiliation{%
  \institution{University of Chicago}
  \country{US}
}
\email{jacy@sentienceinstitute.org}

\renewcommand{\shortauthors}{Manoli, Pauketat, and Anthis}

\begin{abstract}
  Robots and other artificial intelligence (AI) systems are widely perceived as moral agents responsible for their actions. As AI proliferates, these perceptions may become entangled via the \textit{moral spillover} of attitudes towards one AI to attitudes towards other AIs. We tested how the seemingly harmful and immoral actions of an AI or human agent spill over to attitudes towards other AIs or humans in two preregistered experiments. In Study 1 \mbox{(\textit{N} = 720)}, we established the moral spillover effect in human-AI interaction by showing that immoral actions increased attributions of negative moral agency (i.e., acting immorally) and decreased attributions of positive moral agency (i.e., acting morally) and moral patiency (i.e., deserving moral concern) to both the agent (a chatbot or human assistant) and the group to which they belong (all chatbot or human assistants). There was no significant difference in the spillover effects between the AI and human contexts. In Study 2 \mbox{(\textit{N} = 684)}, we tested whether spillover persisted when the agent was individuated with a name and described as an AI or human, rather than specifically as a chatbot or personal assistant. We found that spillover persisted in the AI context but not in the human context, possibly because AIs were perceived as more homogeneous due to their outgroup status relative to humans. This asymmetry suggests a double standard whereby AIs are judged more harshly than humans when one agent morally transgresses. With the proliferation of diverse, autonomous AI systems, HCI research and design should account for the fact that experiences with one AI could easily generalize to perceptions of all AIs and negative HCI outcomes, such as reduced trust.
\end{abstract}

\ccsdesc[500]{Human-centered computing~HCI theory, concepts and models}
\ccsdesc[500]{Human-centered computing~Collaborative and social computing theory, concepts and paradigms}
\ccsdesc[500]{Computing methodologies~Theory of mind}
\ccsdesc[500]{Applied computing~Psychology}

\keywords{Human-AI Interaction, Moral Spillover, Moral Agency, Moral Patiency, Social Psychology}

\maketitle

\section{Introduction}

From surveillance drones to surgical robots, artificial intelligence (AI) with the ability to make autonomous decisions is becoming increasingly widespread. The actions of AIs often have moral consequences. Autonomous vehicles need to make split-second life-and-death decisions \cite{awad18a}. Interaction with chatbots can benefit or harm mental health \cite{lee19d, laestadius22}. Even though AIs currently lack the ability to make conscious moral decisions, the human-computer interaction (HCI) literature has shown that they are readily perceived as moral agents worthy of blame, punishment, and praise for their actions \cite{banks19b, cervantes20a, lima23a}.

However, moral reactions to an AI do not always occur in the same way as they do to a human. For example, \citet{stuart21a} found that people were actually less willing to blame AIs for bad outcomes than for neutral outcomes, arguing that this is because bad outcomes make people attempt to bridge a “responsibility gap" for which they must seek out another agent, such as the human creator, who can be held culpable more easily. Other HCI studies suggest that AIs are blamed more than humans when moral harm occurs \cite{franklin21, hong20a, liu19a, liu21a}, and in non-moral contexts people exhibit a general pattern of “algorithmic aversion” by preferring human advice over that of an algorithm, even when the algorithm has better performance \cite{dietvorst15a, longoni22}.

An important phenomenon in the literature on human-human interaction, which has not yet been studied in human-AI interaction, is how moral attitudes towards one individual generalize to moral attitudes towards other individuals or groups, a phenomenon referred to as moral spillover \cite{mullen08}. For example, \citet{uhlmann12} showed that the biological relatives of a criminal were held morally responsible for the criminal’s actions, suggesting a spillover of attributions of moral agency (i.e., the ability to make moral or immoral choices \cite{banks21, gray09, gray12b}) from the criminal to their relatives. Prosocial interactions with an outgroup member often transfer to more positive evaluations of the outgroup as a whole \cite{boin21, desforges91}. If moral spillover occurs in human-AI interaction, it may be beginning to have significant real-world implications as users interact with many different AI systems in everyday life. Observing a single AI’s seemingly immoral behavior might create negative predispositions towards AIs in general, which could have numerous effects such as reducing trust in AI—even in AIs that are deployed responsibly. In addition to moral agency, spillover could occur for moral patiency (i.e., the ability to be harmed and benefited by others’ actions \cite{banks21, gray09, harris21a, pauketat22b}). While the human-human literature has established the spillover of moral agency, and there are many studies on the attribution of moral patiency to humans and to AI, there have not yet been studies of the effect of an AI’s immoral actions on attribution of moral patiency to the agent itself—an additional gap that we aimed to fill.

\begin{figure}[ht]
    \includegraphics[width=\textwidth]{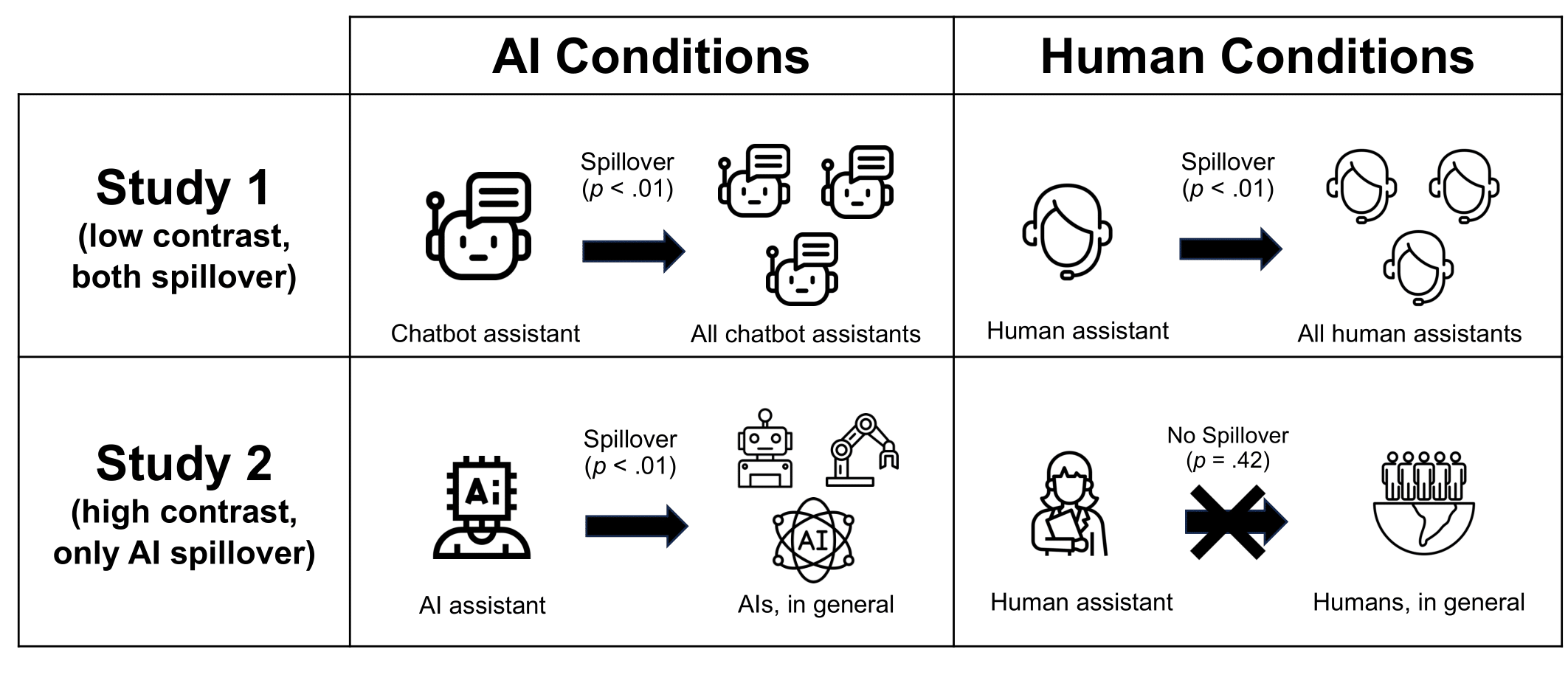}
    \caption{Summary of results. Across two studies, we found that attribution of negative moral agency, positive moral agency, and moral patiency spills over from an AI assistant to a group of similar AI assistants, and also all AIs. Contrarily, moral attribution to a human assistant only spills over to the group of human assistants, but not all humans. (Images adapted from flaticon.com).}
    \label{fig:spillover}
\end{figure}

In this work, we tested whether moral spillover occurs in human-AI interaction and how it compares to human-human interaction in two different scenarios. We assessed attributions of negative moral agency, positive moral agency, and moral patiency. In Study 1, we found that an unnamed chatbot assistant or human personal assistant’s immoral behavior increased attributions of negative moral agency and decreased attributions of positive moral agency and moral patiency to the congruent groups of chatbot or human assistants. In Study 2, we replicated the study in a scenario with a more individuated agent and an expansion of the group from human assistants to all humans and from chatbot assistants to all AIs, which reduced the similarity between the agent and the group. We found that an individuated AI agent's behavior had the same spillover to all AIs, but an individuated human agent's behavior did not affect moral attributions to all humans (see \mbox{\Cref{fig:spillover} for a summary of the results)}. Based on these results, we make the following contributions:

\begin{enumerate}
    \item We demonstrate moral spillover in the AI context. Namely, the immoral action of one AI agent spills over to attributions of moral agency and patiency to the particular group to which the AI belongs and to AIs as a whole. The ease with which spillover can occur suggests that AI systems should be designed in ways that account for the risk that one mistake could taint perceptions of all AIs.
    \item When assessing moral spillover to AIs and humans in general (Study 2), we found a double standard in the asymmetric spillover of moral attributions to AIs and humans. AIs were judged more harshly as a group than humans when one AI acted immorally. This suggests that AIs have a more malleable moral standing that can be affected without explicit similarity between the AI agent and its group. Moral spillover may have larger effects with AIs than it does in the human context.
    \item While we focused on group attributions, in Study 1, the chatbot agent itself was attributed more negative moral agency and was judged more harshly as an individual than was the human agent. This was in line with the current literature, but in Study 2, the AI agent was actually attributed less negative moral agency and judged less harshly as an individual than was the human agent. This deviation from past findings suggests an important boundary condition in which reactions to individuated AIs of a more general category may different from reactions to non-individuated and specialized AIs.
\end{enumerate}

\section{Related work}

Even though the moral decisions of an AI agent directly affect how the agent is perceived, it is currently unknown whether and how these perceptions might affect other AIs. This is essential to grapple with as AI systems become more common and take on more complex roles in society. Because these are the first studies of moral spillover in the AI context, we ground our work in the literature on moral attributions to individual AI agents and in the literature on moral spillover in the human-human context.

\subsection{Moral agency and patiency attributions to AI}

While it has been shown that humans attribute moral agency and responsibility to AIs, it is not clear whether AIs are attributed more or less than that attributed to other humans. AIs have been blamed and held responsible more than humans for accidents in the case of self-driving cars \cite{franklin21, hong20a, liu19a, liu21a} and when failing to intervene in moral dilemmas where people’s lives could be saved \cite{komatsu21a}. This could be partly caused by “automation bias” where humans have higher expectations for AI to be useful and competent in HCI settings, and therefore people judge mistakes or transgressions made by AI more harshly \cite{alon-barkat23, rebitschek21}. Nevertheless, there are cases in which AIs are also blamed less than humans. People seem to be outraged more at humans who discriminate than AIs who discriminate \cite{bigman23}, and service robots were shown to be blamed less than the firms deploying them when making mistakes \cite{leo20, ryoo24}. This is consistent with Stuart and Kneer’s \cite{stuart21a} argument that people seek to assign blame for harmful actions to humans rather than AIs. In this study, 513 participants recruited from Amazon Mechanical Turk (MTurk) read vignettes in which the decisions of a human agent, an AI agent, or a corporation put others at risk of harm. Results showed that the AI agent was blamed less than the other agents when the outcome of the decision caused harm. Some of these effects were associated with AIs’ perceived control over a situation, which is often characterized as less than a human’s \cite{leo20, ryoo24}. 

There is a quickly growing literature on the attribution of moral patiency to AI. \citet{lima20a} and \citet{pauketat22b} demonstrated that people are willing to consider AIs as moral patients. Lima et al. tested a range of interventions that could increase the attribution of patiency—namely the support of AI rights—such as sharing information about the rights of other nonhuman entities. Pauketat and Anthis found that mind perception of current AIs was positively correlated with support for AI rights and mind perception of future AIs. In the context of the present work, attributions of moral patiency to an agent may be affected by their immoral actions, but that effect has not yet been shown in the human-AI literature. However, in the human-human literature, \citet{yu23} showed that people cared less about the welfare of humans who behaved immorally, such as people who physically harmed or who discriminated against coworkers. Across four vignette-based experiments, 421 participants recruited from Prolific withheld their sympathy for immoral people in need of help, even when their immoral actions were not related to the cause of their suffering.

\subsection{Moral spillover}

While there have not yet been studies on moral spillover of attitudes towards AI, the presence of other dynamics in HCI suggests its importance. The aforementioned algorithmic aversion and automation bias have been explored in terms of spillover by \citet{longoni23}. The authors used 13 vignette-based experiments with over 3,000 participants recruited from Amazon MTurk to examine how an AI's wrong decisions in the context of government welfare programs transfers to evaluations of other AIs with similar responsibilities. They found that an AI’s failure generalized more to evaluations of other AIs than a human’s failure generalized to other humans, which they term “algorithmic transference.” Within the moral domain, \citet{chernyak16a} demonstrated that a child with a pet dog in their household treated robot dogs better than children without pet dogs, which may be a result of the spillover of moral consideration from nonhuman animals to animal-like robots. 

It is currently unknown how moral attributions generalize from an AI agent to groups of AIs with varying degrees of similarity to the agent. Research in human-human interaction suggests that a minimum threshold of similarity (e.g., kinship) must be reached for spillover to occur \cite{boin21, desforges91, uhlmann12}. In the case of AIs, all of them may be perceived as a homogeneous group and therefore reach this threshold, such that the moral transgression of an AI signals that all AIs are comparably immoral even without any explicit similarity. Differences in spillover like this may be expected given the generally divergent attributions of responsibility to human and AI agents \cite[e.g.,][]{franklin21, hong20a, ryoo24}. For example, Hong examined blame attribution in the case of self-driving cars versus human drivers. In this study, 248 participants recruited from Amazon MTurk read hypothetical vignettes describing car accidents of varying severity (i.e., the victim survived or died). The self-driving car was consistently blamed more than the human, with accident severity exacerbating blame attribution. Such asymmetries between AIs and humans in moral spillover could reflect a more malleable moral standing of AIs in society and a greater difficulty of forming nuanced moral attitudes and beliefs that distinguish one AI from many AIs, increasing the effects of particular interactions on evaluations of all AIs.
 
In human-AI interaction broadly, a spillover of negative moral attributions from an individual AI to all AIs could lead to mistreatment \cite{arrambide22a, freier08a, kahn12a, ladak23d, ladak23e} or a lack of trust \cite{dietvorst15a, hatherley20, longoni22}; be detrimental for humans’ overall social and moral behavior \cite{whitby08a}; or limit the harmonious relationships that humans could build with AIs \cite[e.g.,][]{berberich20, do23a} by exacerbating the perception of AIs as worthless, dangerous, unreliable, or incompetent.

Based on this literature, we designed two experiments to examine moral spillover in human-AI interaction. We based the design of these experiments on previous HCI studies that used vignettes to test responses to hypothetical human-AI scenarios \mbox{\cite{stuart21a, hong20a, longoni23, yu23}}. In Study 1, we tested the spillover of three attitudes (positive and negative moral agency and moral patiency) toward a single chatbot or human personal assistant to attitudes toward chatbots or human personal assistants in general. In Study 2, we followed the same overall methodology but with less similarity between the agent and their group by giving the agent a name and labeling the group as "AIs, in general" or "humans, in general" rather than the narrower category of assistants. We propose and test the following hypotheses for Study 1 and Study 2:

\begin{itemize}
    \item \textbf{Agent effects (\textit{H1})}: A human or AI agent’s immoral action (a) increases attribution of negative moral agency to the agent, (b) decreases attribution of positive moral agency to the agent, and (c) decreases attribution of moral patiency to the agent, relative to the morally neutral action.
    \item \textbf{Spillover effects (\textit{H2})}: A human or AI agent’s immoral action (a) increases attribution of negative moral agency to the group the agent belongs to (humans or AIs), (b) decreases attribution of positive moral agency to the group the agent belongs to (humans or AIs), and (c) decreases attribution of moral patiency to the group the agent belongs to (humans or AIs), relative to the agent’s morally neutral action.
    \item \textbf{Spillover asymmetry (\textit{H3})}: An AI agent's immoral action spills over to attributions of (a) negative moral agency, (b) positive moral agency, and (c) moral patiency to the AI group differently than a human agent's immoral action spills over to attributions to the human group.
\end{itemize}

\section{Experiments}

\begin{figure}[ht]
    \includegraphics[scale=0.5]{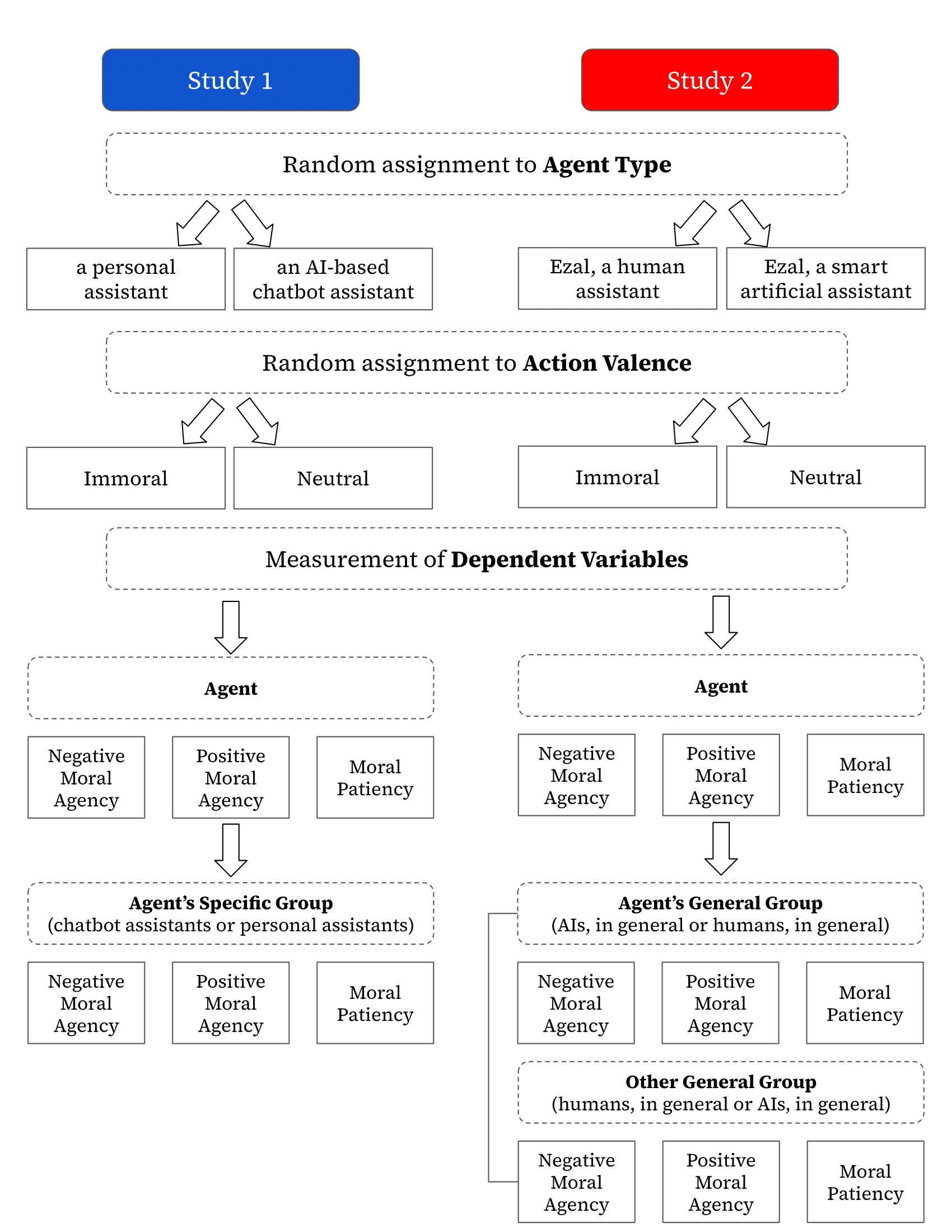}
    \caption{Flowchart of Study 1 and Study 2 procedure. In Study 1, participants were randomly assigned to read a vignette about a chatbot or human personal assistant who acted immorally or neutrally in their workplace, and rated the assistant and their specific group (chatbots or personal assistants) on negative moral agency, positive moral agency, and moral patiency. In Study 2, the similarity between agent and group was decreased by individuating the agent with a name ("Ezal") and measuring moral attribution of the agent's general group (humans or AIs, in general).}
    \label{fig:flowchart}
\end{figure}

To test our hypotheses, we conducted two online vignette experiments that were preregistered on AsPredicted (Study 1: \url{https://aspredicted.org/VFD_MM3}; Study 2: \url{https://aspredicted.org/KFY_Q8C}). Both studies tested all hypotheses. Study 1 investigated spillover from the actions of  a single hypothetical AI chatbot or human personal assistant's actions in the workplace to attitudes towards those groups. Study 2 used the same vignette scenario but individuated the assistant with a name, had the groups of AIs and humans in general rather than assistants in particular, and asked participants to evaluate both AI and human groups rather than only the group matching the agent in the treatment condition (see \mbox{\Cref{fig:flowchart}}). All materials, data, and analysis code are available on OSF (\url{https://osf.io/s4gvy/?view_only=7f60b63e04724b7c88eeaf85dd88cfc8}). All procedures were in accordance with the 1964 Helsinki declaration and its later amendments. Participants gave their informed consent prior to and were debriefed after each study.

\subsection{Study 1}

In Study 1, we assessed the spillover of moral attributions from a hypothetical unnamed chatbot assistant or human personal assistant to chatbot assistants or human personal assistants in general. We aimed to establish whether spillover occurs between an AI agent and a group that includes the agent and other entities high in similarity, and to compare it to spillover in an identical context with a human agent and group. To do this, we created a scenario in which the assistant performs an immoral or morally neutral action in their workplace. Participants rated the moral agency and patiency of the agent and the group congruent to the agent type they read about.

\subsubsection{Participants}

A G*Power analysis, a commonly used software for power analysis in the social and behavioral sciences \mbox{\cite{faul07, kang21}}, showed that 787 participants were needed to detect a small effect size (Cohen’s $f$ = .10) with 80\% power and $\alpha$ = .05. We recruited 866 U.S. online participants from Prolific to account for possible attrition \cite{hochheimer16}. Of those, 139 participants were excluded from analyses due to incomplete responses or failing the attention and reading comprehension checks. The final sample was 727 participants ($M_{age}$ = 42.23, $SD_{age}$ = 13.79, 50\% female, 66\% White), providing 97\% power with post-hoc effect size estimation. Each participant was randomly assigned to one of the four conditions (immoral chatbot assistant: $N$ = 181; morally neutral chatbot assistant: $N$ = 193; immoral human assistant: $N$ = 186; morally neutral human assistant: $N$ = 160).

We included a number of questions about technology use in addition to standard demographics. Most participants, 89.6\%, owned smartphones; 47.5\% reported owning AI or robotic devices; and 26.8\% reported using AI or robotic devices in their workplace. Participants reported a moderate level of exposure to AIs via direct interaction or narratives in various media (direct interaction: $M$ = 2.32, $SD$ = 1.90; AI narratives: $M$ = 2.11, $SD$ = 1.40; each on a scale of 0 = never to 5 = daily).

\subsubsection{Materials and procedure}

\subparagraph{\hspace{1em}\textit{Experimental manipulations.}} 

Participants read a vignette describing the workplace tasks of an individual who was either described as “a personal assistant” or “an AI-based chatbot assistant.” A definition of chatbot and human assistants was included to ensure clarity, particularly to reinforce the fact that the assistant was an AI or was a human (see Supplementary Materials). In the vignette, which was designed to be as neutral as possible and describe actions that a human or AI could plausibly do with minimal variation between treatment conditions, the agent “does math calculations, solves problems, searches for information, and summarizes information.” Participants then read about a day at work in which the agent’s department had an important deadline. In the immoral action condition, the agent intentionally downloaded malicious software on company-owned equipment in order to use the damage caused as an excuse to fall behind with their tasks. Because of this, they heavily disrupted the work of their coworkers, and many employees were scolded for poor performance. In the morally neutral action condition, the agent followed standard procedure to download specialized software on company-owned equipment in order to complete their tasks alongside their coworkers (without an explicit intention to help or harm anyone). In this case, the agent’s work was standard, and their coworkers continued their usual work without an explicitly negative or positive outcome resulting from the agent's action, making this as a neutral action.  

\paragraph{Comprehension checks}

All participants responded to two multiple choice questions about the content of the vignettes. Participants were asked how the agent acted at work (e.g., “Harmed their coworkers’ work,” “Followed standard procedure at work”) and what kind of entity the agent was (e.g., “AI,” “human”).

\paragraph{Dependent variables}

Participants responded to questions for each of the six dependent variables (randomly presented): negative moral agency, positive moral agency, and moral patiency for the agent and either human assistants in general (if they read about a human) or chatbot assistants in general (if they read about a chatbot).

\begin{itemize}
    \item \textbf{Moral agency} was measured with composite scales based on the six dimensions of Moral Foundations Theory \cite[MFT;][]{graham11, iyer12}: Care/Harm, Fairness/Cheating, Loyalty/Betrayal, Authority/Subversion, Sanctity/Degradation, and Liberty/Oppression. A similar approach was used in a previous study on attributions of moral agency to AIs \cite{banks21a}. We created one positive and one negative item for each of the six MFT dimensions (12 items in total), corresponding to moral and immoral actions, respectively. For example, participants evaluated how likely the human or AI agent was to “save someone in danger even if they might get hurt in the process” (Care/Harm; positive) and to “do something illegal even if an authority figure told them not to” (Authority/Subversion; negative). All items were measured on a continuous scale from 1 (not at all) to 7 (very much). Positive and negative items were averaged to produce composite positive and negative moral agency scores. The subscales demonstrated good internal consistency for the human assistant (positive: Cronbach’s $\alpha$ = .87; negative: $\alpha$ = .94), the chatbot assistant (positive: $\alpha$ = .86; negative: $\alpha$ = .94), human assistants (positive: $\alpha$ = .87; negative: $\alpha$ = .92), and chatbot assistants (positive: $\alpha$ = .87; negative: $\alpha$ = .92). The subscales also demonstrated high correlations with another moral agency scale (Banks, 2019) for the human assistant (positive: Pearson’s $r$(725) = .64, $p$ < .001; negative: $r$(725) = -.58, $p$ < .001), the chatbot assistant (positive: $r$(725) = .65, $p$ < .001; negative: $r$(725) = -.56, $p$ < .001), human assistants (positive: $r$(386) = .51, $p$ < .001; negative: $r$(386) = -.33, $p$ < .001), and chatbot assistants (positive: $r$(436) = .52, $p$ < .001; negative: $r$(436) = -.34, $p$ < .001).
    \item \textbf{Moral patiency} was measured with six items adapted from the AI treatment scale \cite[e.g., “Robots/AIs deserve to be treated with respect”][]{pauketat21}, measured on a continuous scale from 1 (strongly disagree) to 7 (strongly agree). All items were averaged together for the agent and averaged together for the group. This scale also demonstrated good internal consistency (human assistant: $\alpha$ = .89; chatbot assistant: $\alpha$ = .88; human assistants: $\alpha$ = .88; chatbot assistants: $\alpha$ = .87).
\end{itemize}

\paragraph{Other measures}

For comparison, we collected an explicit measure of moral character attributed to the agent and the group (e.g., “How moral or immoral are chatbot assistants, in general?”) from 0 (very immoral) to 100 (very moral). We also measured individual differences in AI literacy, the tendency to anthropomorphize, belief in the possibility of AI sentience, and belief in the possibility of AI moral agency. These measures were used to account for the possibility that individual differences in tendency to attribute relevant features may moderate or be associated with attributions of moral agency and patiency. The tendency to anthropomorphize ($\alpha$ = .76) was the averaged responses to eight items (e.g., “To what extent does the wind have intentions?”) on a continuous scale (0 = strongly disagree, 7 = strongly agree), using the interval originally developed for this measure in previous work, which, like 0 to 100 for non-agreement measures, is commonly used in the behavioral sciences \mbox{\cite{waytz14}}. AI literacy ($\alpha$ = .58) was the averaged responses to twelve items (e.g., “I can distinguish between smart devices and non-smart devices”) on a continuous scale (0 = strongly disagree, 10 = strongly agree) —again, using the interval from the work that originally developed this item \mbox{\cite{wang23c}}. Belief in AI sentience and moral agency were measured with one yes/no question each: “Do you think it could ever be possible for AIs to be [sentient/moral agents]?” We also measured how much participants liked the agent on a continuous scale from 0 (not at all) to 100 (very much) to control for the possibility that likeability towards the agent explains moral attributions. Lastly, we measured standard demographic characteristics, namely age, gender, ethnicity, annual income, and education level. These measures were presented in random order. Results that included all abovementioned covariates in the models are in the Supplementary Materials.

\subsubsection{Data analysis}

To test our hypotheses, we performed 2x2 between-subjects analyses of variance (ANOVAs) in R 4.3.1. Attributions of negative moral agency, positive moral agency, and moral patiency were tested in separate models with agent type (human, AI) and action valence (immoral, morally neutral) as between-subjects independent variables. The results reported below were consistent with additional analyses controlling for the false discovery rate \cite{benjamini95a} and covariates (see Supplementary Materials).

\subsubsection{Agent effects (\textit{H1})}

\begin{quote}
    \textit{A human or AI agent’s immoral action (a) increases attribution of negative moral agency, (b) decreases attribution of positive moral agency, and (c) decreases attribution of moral patiency to the agent, relative to the morally neutral action.}
\end{quote}

\paragraph{Agent’s negative moral agency}

\begin{figure}[ht]
    \includegraphics[width=\textwidth]{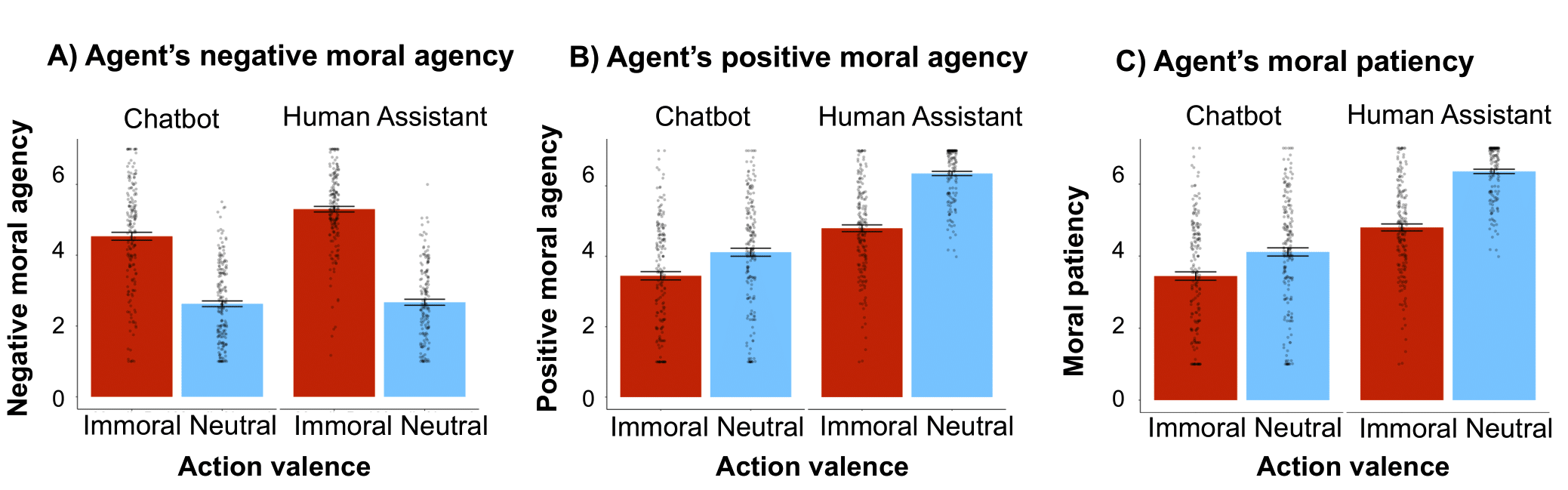}
    \caption{Effect of agent type and action valence on the agent’s moral attributions. Bars represent the mean and error bars represent the standard error of the mean. Dots represent individual data points. A) Negative moral agency increased in the immoral condition compared to the morally neutral condition, more so for the chatbot than the human assistant. B) Positive moral agency decreased in the immoral condition compared to the morally neutral condition, more so for the human than the chatbot assistant. C) Moral patiency decreased in the immoral condition compared to the morally neutral condition, more so for the human than the chatbot assistant.}
    \label{fig:s1agent}
    \phantomsubcaption\label{fig:s1agentA}
    \phantomsubcaption\label{fig:s1agentB}
    \phantomsubcaption\label{fig:s1agentC}
\end{figure}

In the 2x2 model with agent type and action valence, there was a significant main effect of action valence on negative moral agency, supporting \textit{H1a}. There was also a significant effect of agent type and an agent type-valence interaction. Post-hoc Tukey’s honest significant difference (HSD) tests showed that negative moral agency increased in the immoral compared to the morally neutral condition for chatbot-assigned participants (immoral: $M$ = 4.53, SE = 0.11; morally neutral: $M$ = 2.63, SE = 0.08, $p$ < .001) and for human-assigned participants (immoral: $M$ = 5.30, SE = 0.08; morally neutral: $M$ = 2.67, SE = 0.09, $p$ < .001), supporting \textit{H1a} for each group. There was also a significant difference between the group-specific effects of valence, such that negative moral agency increased more for the chatbot than the human agent ($p$ < .001). (\Cref{tab:s1table1} and \Cref{fig:s1agentA})

\begin{table}[ht]
\centering
\caption{Study 1 2x2 ANOVA results for negative moral agency, positive moral agency, and moral patiency for the chatbot or human assistant
}
\begin{tabular}{lllllll}
\hline
Dependent Variable    & Independent Variable & \textit{F} & \textit{df1} & \textit{df2} & \textit{p}      & \textit{$\eta^2_p$} \\
\hline
Negative moral agency & Valence              & 616.50     & 1            & 716          & \textless .001* & .28                                 \\
                      & Agent type           & 34.70      & 1            & 716          & \textless .001* & .05                                 \\
                      & Valence x agent type & 15.90      & 1            & 716          & \textless .001* & .02                                 \\
Positive moral agency & Valence              & 274.97     & 1            & 716          & \textless .001* & .28                                 \\
                      & Agent type           & 0.98       & 1            & 716          & .32             & .00                                 \\
                      & Valence x agent type & 49.92      & 1            & 716          & \textless .001* & .07                                 \\
Moral patiency        & Valence              & 115.20     & 1            & 716          & \textless .001* & .14                                 \\
                      & Agent type           & 286.20     & 1            & 716          & \textless .001* & .29                                 \\
                      & Valence x agent type & 18.80      & 1            & 716          & .03*            & .03                                \\
\hline
\end{tabular}
\tablefootnote{* Significant \textit{p}-values (\textit{p} < .05)}
\label{tab:s1table1}
\end{table}

\paragraph{Agent’s positive moral agency}

There was a significant main effect of action valence on the agent’s perceived positive moral agency, supporting \textit{H1b}. There was also a significant effect of agent type-valence interaction. Post-hoc Tukey’s HSD tests showed that positive moral agency decreased in the immoral compared to the morally neutral condition for chatbot-assigned participants (immoral: $M$ = 3.58, SE = 0.10; morally neutral: $M$ = 4.45, SE = 0.09, $p$ < .001) and human-assigned participants (immoral: $M$ = 2.96, SE = 0.08; morally neutral: $M$ = 5.08, SE = 0.06, $p$ < .001), supporting \textit{H1b} for each group. There was also a significant difference between the group-specific effects of valence, such that positive moral agency decreased more for the human than the chatbot agent ($p$ < .001). The main effect of agent type was not significant. (\Cref{tab:s1table1} and \Cref{fig:s1agentB})

\paragraph{Agent’s moral patiency}

There was a significant main effect of action valence on the agent’s moral patiency, supporting \textit{H1c}. There was also a significant effect of agent type and agent type-valence interaction. Post-hoc Tukey’s HSD tests showed that moral patiency decreased in the immoral compared to the morally neutral condition for chatbot-assigned participants (immoral: $M$ = 3.45, SE = 0.11; morally neutral: $M$ = 3.23, SE = 0.11, $p$ < .001) and human-assigned participants (immoral: $M$ = 5.53, SE = 0.07; morally neutral: $M$ = 5.89, SE = 0.06), supporting \textit{H1c} for each group. There was also a significant difference between the group-specific effects of valence, such that moral patiency decreased more for the human than the chatbot agent ($p$ < .001). (\Cref{tab:s1table1} and \Cref{fig:s1agentC})

\subsubsection{Spillover effects (\textit{H2}) and spillover asymmetry (\textit{H3})}

\begin{quote}
    \textit{A human or AI agent’s immoral action (a) increases attribution of negative moral agency, (b) decreases attribution of positive moral agency, and (c) decreases attribution of moral patiency to the group the agent belongs to (humans or AIs), relative to the agent’s morally neutral action (\textit{H2}). An AI agent's immoral action spills over to attributions of (a) negative moral agency, (b) positive moral agency, and (c) moral patiency to the AI group differently than a human agent's immoral action spills over to attributions to the human group (\textit{H3}).}
\end{quote}

\paragraph{Group’s negative moral agency}

\begin{figure}[ht]
    \includegraphics[width=\textwidth]{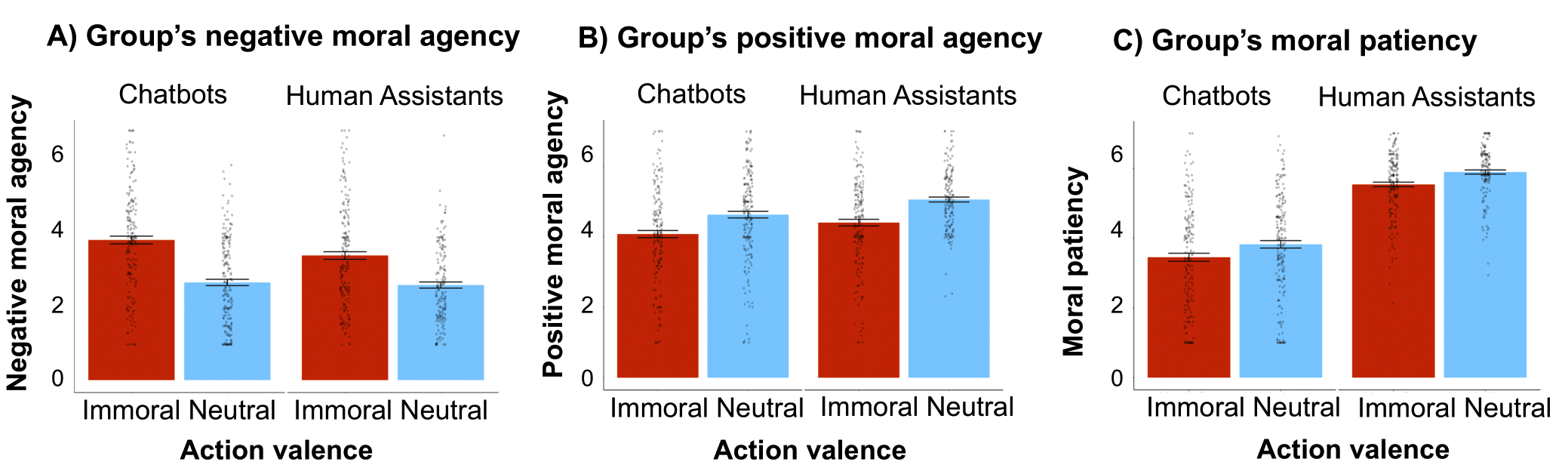}
    \caption{Effect of agent type and action valence on the moral attributions to chatbot and human assistants in general. Bars represent the mean and error bars represent the standard error of the mean. Dots represent individual data points. A) Negative moral agency increased in the immoral condition compared to the morally neutral condition. B) Positive moral agency decreased in the immoral condition compared to the morally neutral condition. C) Moral patiency decreased in the immoral condition compared to the morally neutral condition.}
    \label{fig:s1group}
    \phantomsubcaption\label{fig:s1groupA}
    \phantomsubcaption\label{fig:s1groupB}
    \phantomsubcaption\label{fig:s1groupC}
\end{figure}

In the 2x2 model between agent type and action valence, there was a significant main effect of action valence on negative moral agency attributions to the group the agent belongs to (human or chatbot assistants), supporting \textit{H2a}. Post-hoc Tukey’s HSD tests showed that negative moral agency increased in the immoral compared to the morally neutral condition for chatbot assistants (immoral: $M$ = 3.93, SE = 0.11, morally neutral: $M$ = 2.74, SE = 0.09, $p$ < .001) and human assistants (immoral: $M$ = 3.49, SE = 0.09; morally neutral: $M$ = 2.67, SE = 0.09, $p$ < .001), supporting \textit{H2a} for each group. The agent type-valence interaction was not significant, which opposes \textit{H3a} because the immoral action increased negative moral agency relative to the neutral action in a similar way for both chatbot assistants and human assistants, meaning that there was no asymmetry in the spillover of negative moral agency between the groups. Lastly, there was also a significant effect of agent type. Chatbots had significantly higher overall negative moral agency than human assistants (chatbots: $M$ = 3.32, SE = 0.08; human assistants: $M$ = 3.11, SE = 0.07). (\Cref{tab:s1table2} and \Cref{fig:s1groupA})

\begin{table}[ht]
\centering
\caption{Study 1 2x2 ANOVA results for negative moral agency, positive moral agency, and moral patiency for chatbots or human assistants
}
\begin{tabular}{lllllll}
\hline
Dependent Variable    & Independent Variable & \textit{F} & \textit{df1} & \textit{df2} & \textit{p}      & \textit{$\eta^2_p$} \\
\hline
Negative moral agency & Valence              & 105.75     & 1            & 716          & \textless .001* & .13                                 \\
                      & Agent type           & 4.24       & 1            & 716          & .04*            & .00                                 \\
                      & Valence x agent type & 3.38       & 1            & 716          & .07             & .00                                 \\
Positive moral agency & Valence              & 42.82      & 1            & 716          & \textless .001* & .06                                 \\
                      & Agent type           & 13.99      & 1            & 716          & \textless .001* & .02                                 \\
                      & Valence x agent type & 0.31       & 1            & 716          & .58             & .00                                 \\
Moral patiency        & Valence              & 15.61      & 1            & 716          & \textless .001* & .02                                 \\
                      & Agent type           & 502.25     & 1            & 716          & \textless .001* & .41                                 \\
                      & Valence x agent type & 0.01       & 1            & 716          & .94             & .00       
                      \\
\hline
\end{tabular}
\tablefootnote{* Significant \textit{p}-values (\textit{p} < .05)}
\label{tab:s1table2}
\end{table}

\paragraph{Group’s positive moral agency}

There was a significant main effect of action valence on positive moral agency attributions to the congruent group, supporting \textit{H2b}. Post-hoc Tukey’s HSD tests showed that positive moral agency decreased in the immoral compared to the morally neutral condition for chatbot assistants (immoral: $M$ = 4.08, SE = 0.01, morally neutral: $M$ = 5.05, SE = 0.07, $p$ < .001) and human assistants (immoral: $M$ = 4.40, SE = 0.09; morally neutral: $M$ = 4.63, SE = 0.09, $p$ < .001), supporting \textit{H2b} for each group. The agent type-valence interaction was not significant. This opposes \textit{H3b} because the immoral action decreased positive moral agency relative to the neutral action in a similar way for both chatbot assistants and human assistants, meaning that there was no asymmetry in the spillover of positive moral agency between the groups. Lastly, there was also a significant effect of agent type. Human assistants also had significantly higher overall positive moral agency than chatbots (human assistants: $M$ = 4.71, SE = 0.06; chatbots: $M$ = 4.36, SE = 0.07). (\Cref{tab:s1table2} and \Cref{fig:s1groupB})

\paragraph{Group’s moral patiency}

There was a significant main effect of action valence on moral patiency attributions to the congruent group, supporting \textit{H2c}. Post-hoc Tukey’s HSD tests showed that positive moral agency decreased in the immoral compared to the morally neutral condition for chatbot assistants (immoral: $M$ = 3.45, SE = 0.11, morally neutral: $M$ = 3.82, SE = 0.11, $p$ < .001) and human assistants (immoral: $M$ = 5.53, SE = 0.07; morally neutral: $M$ = 5.89, SE = 0.06, $p$ < .001), supporting \textit{H2c} for each group. The agent type-valence interaction was not significant. This opposes \textit{H3c} because the immoral action decreased moral patiency relative to the neutral action in a similar way for both chatbot assistants and human assistants, meaning that there was no asymmetry in the spillover of moral patiency between the groups. Lastly, there was a significant effect of agent type. Human assistants had significantly higher overall moral patiency than chatbots (human assistants: $M$ = 5.70, SE = 0.05; chatbots: $M$ = 3.64, SE = 0.08). (\Cref{tab:s1table2} and \Cref{fig:s1groupC})

\subsubsection{Discussion}

In Study 1, we found that the immoral action of a chatbot or human assistant increased attribution of negative moral agency and decreased attribution of positive moral agency and moral patiency to the agent relative to a neutral action. These attributions spilled over to the group the agent belongs to, chatbot or human assistants. Our findings support \textit{H1} and \textit{H2}. The spillover of moral attributions from one AI to a group of similar AIs extends previous findings that demonstrated generalization between singular AI agents in HCI \cite{chernyak16a, longoni23} into a moral context. This highlights the importance of the actions of one AI for the perception of other AIs. If one AI morally transgresses, this could have far-reaching effects.

However, we found no asymmetry between the spillover of moral attributions to chatbot assistants and human personal assistants, contrary to \textit{H3} and suggesting that spillover may occur in both human and AI contexts. Our findings are in tension with \citet{longoni23}, where algorithmic failure generalized more than human failure in the context of AI-assisted government welfare programs, but they parallel \citet{uhlmann12}, where attributions of moral agency generalized between human agents who were biologically similar to a human criminal. In our scenario, human assistants may have reached a crucial threshold of similarity to the human assistant, triggering a spillover of moral attributions to the group comparable to the spillover observed between similar AIs. This could be underpinned by people's perception of personal assistants as a relatively specific, homogeneous category, thus readily generalizing the action of one personal assistant to all of them. This is in line with the well-established phenomenon of outgroup similarity, where outgroup members are perceived as more similar to each other and less diverse than ingroup members \mbox{\cite{linville97, ostrom92, sedikides93}}. In our study, both chatbot and personal assistant groups were relatively specific and could have been perceived as homogeneous, leading to comparable spillover to both groups.

\subsection{Study 2}

In Study 2, we tested whether moral spillover is observed when the agent is individuated with a name, the scope is expanded from assistants to all AIs and all humans, and participants are aware that both humans and AI agents are being evaluated in the between-subjects experiment. These three changes reduced the similarity (i.e., increased the contrast) between the agent and the group, relative to Study 1.

\subsubsection{Participants}

The methodology of Study 2 largely mirrored that of Study 1. We recruited 866 U.S. participants online from Prolific to detect a small effect size (Cohen’s $f$ = .10; 80\% power; $\alpha$ = .05) and account for possible attrition. Of those, 182 participants were excluded from analyses due to incomplete responses or failing the attention and reading comprehension checks. The final sample was 684 participants ($M_{age}$ = 41.30, $SD_{age}$ = 13.70, 50\% female, 70\% White), providing 75\% power. Each participant was randomly assigned to one of the four conditions (immoral AI: $N$ = 179; morally neutral AI: $N$ = 183; immoral human: $N$ = 173; morally neutral human: $N$ = 149).

Most participants, 86.9\%, owned smartphones; 47.8\% owned AI or robotic devices; and 20.9\% reported using AI or robotic devices in their workplace. Participants reported a moderate level of exposure to robots or AIs via direct interaction or AI narratives in various media (direct interaction: $M$ = 2.26, $SD$ = 1.96; AI narratives: $M$ = 2.44, $SD$ = 1.46; variables measured on a scale of 0 = never to 5 = daily).

\subsubsection{Materials and procedure}

\subparagraph{\hspace{1em}\textit{Experimental manipulations.}} 

Participants read a vignette describing the workplace tasks of an individual named Ezal, either described as “a human assistant” or “a smart artificial assistant” in the human or AI conditions. Unlike Study 1, we did not define “assistant.” The name “Ezal" was chosen based on a pilot study ($N$ = 124) in which it evoked middling perceptions of human-likeness, consciousness, and pleasantness, between strongly human-like (e.g., “Jordan”) and strongly AI-like (e.g., “XZ103”) names. Other than the name and label, the vignettes remained the same as in Study 1.

\paragraph{Comprehension checks}

All participants responded to the same two reading comprehension questions as in Study 1.

\paragraph{Dependent variables}

Participants responded to questions for each of the nine dependent variables (randomly presented): positive moral agency, negative moral agency, and moral patiency for the agent, humans in general, and AIs in general. To allow for additional analysis, unlike Study 1, each participant answered questions regarding both groups. This means that participants were aware that humans were under consideration in the AI condition and that AIs were under consideration in the human condition, further decreasing agent-group similarity. The moral attribution variables were the same scales used in Study 1. The positive and negative moral agency scales demonstrated good internal consistency for the agent (positive: Cronbach’s $\alpha$ = .88; negative: $\alpha$ = .95), humans (positive: $\alpha$ = .81; negative: $\alpha$ = .90), and AIs (positive: $\alpha$ = .81; negative: $\alpha$ = .90). The moral patiency scale also demonstrated good internal consistency (agent: $\alpha$ = .89; humans: $\alpha$ = .84; AIs: $\alpha$ = .86).

\paragraph{Other measures}

As in Study 1, we measured moral character for the agent, humans, and AIs, individual differences in the tendency to anthropomorphize ($\alpha$ = .74), belief in AI sentience and belief in AI moral agency, the agent’s likability, as well as additional demographic characteristics. These measures were all presented in random order. Results that included these covariates in the models are in the Supplementary Materials. Note that we did not measure AI literacy, which had low internal consistency in Study 1 and a lack of significant effects in the supplementary models.

\subsubsection{Data analysis}

As in Study 1, we performed 2x2 between-subjects ANOVAs with agent type (human, AI) and action valence (immoral, morally neutral) as between-subjects independent variables to identify attributions of moral agency and patiency to the agent, AIs in general, and humans in general.

We conducted an additional analysis using the cross-group questions (i.e., AI attributions for human-assigned participants and human attributions for AI-assigned participants). By controlling for responses to these, we were able to test specifically for the spillover from agent to group. For example, in Study 1 the human assistant’s immoral action might have led participants to hold a more general negative worldview that provoked negative assessments of AIs, humans, and other beings, while the chatbot assistant’s immoral action led to a specific negative view towards AI. Potential differences like this were not identifiable in Study 1. To identify them, we conducted 2x2x2 mixed ANOVAs with group type (humans and AIs) as an additional within-subjects variable. The results were consistent with the results reported below, and this analysis is in the Supplementary Materials. The results reported below were also consistent with additional analyses controlling for the false discovery rate \cite{benjamini95a} and covariates (see Supplementary Materials).

\subsubsection{Agent effects (\textit{H1})}

\begin{quote}
    \textit{A human or AI agent’s immoral action (a) increases attribution of negative moral agency, (b) decreases attribution of positive moral agency, and (c) decreases attribution of moral patiency to the agent, relative to the morally neutral action.}
\end{quote}

\paragraph{Agent’s negative moral agency}

\begin{figure}[ht]
    \includegraphics[width=\textwidth]{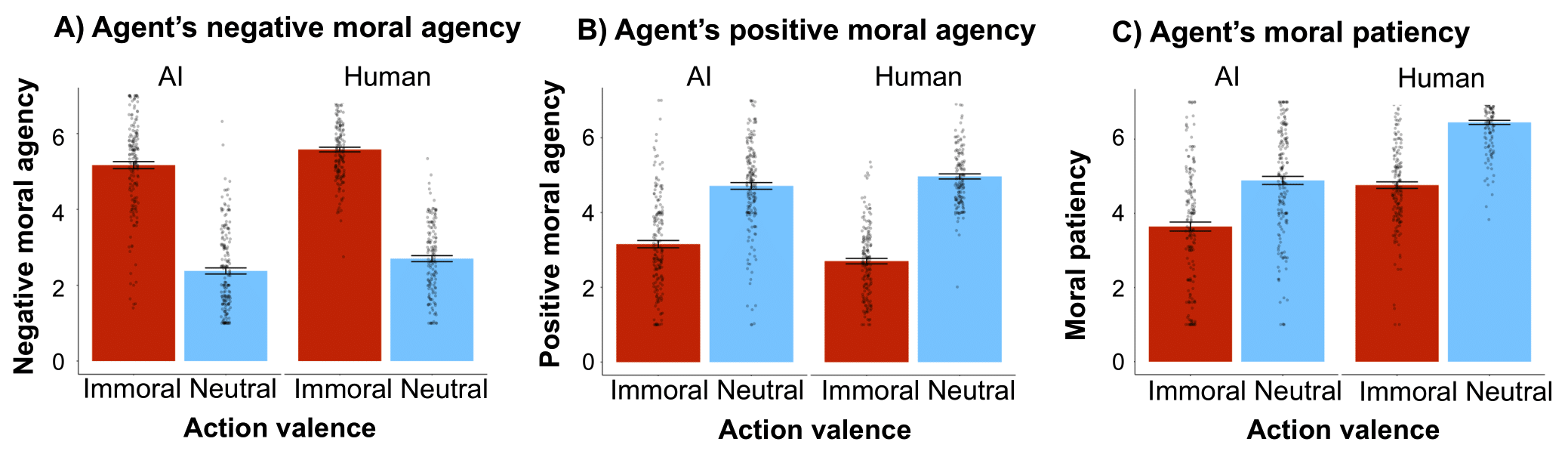}
    \caption{Effect of agent type and action valence on the agent’s moral attributions. Bars represent the mean and error bars represent the standard error of the mean. Dots represent individual data points. A) Negative moral agency increased in the immoral condition compared to the morally neutral condition, more so for the human than the AI agent. B) Positive moral agency decreased in the immoral condition compared to the morally neutral condition, more so for the human than the AI agent. C) Moral patiency decreased in the immoral condition compared to the morally neutral condition for both agent types, more so for the human than the AI agent.}
    \label{fig:s2agent}
    \phantomsubcaption\label{fig:s2agentA}
    \phantomsubcaption\label{fig:s2agentB}
    \phantomsubcaption\label{fig:s2agentC}
\end{figure}

In the 2x2 model between agent type and action valence, there was a significant main effect of action valence on negative moral agency, supporting \textit{H1a}. Post-hoc Tukey’s HSD tests showed that negative moral agency increased in the immoral compared to the morally neutral condition for AI-assigned participants (immoral: $M$ = 5.17, SE = 0.09, morally neutral: $M$ = 2.38, SE = 0.08, $p$ < .001) and human-assigned participants (immoral: $M$ = 5.58, SE = 0.06; morally neutral: $M$ = 2.70, SE = 0.08, $p$ < .001), supporting \textit{H1a} for each group.There was also a significant effect of agent type, such that the human agent had higher negative moral agency ($M$ = 4.25, SE = 0.09) than the AI agent ($M$ = 3.76, SE = 0.10). The agent type-valence interaction was not significant. (\Cref{tab:s2table1} and \Cref{fig:s2agentA})

\begin{table}[ht]
\centering
\caption{Study 2 2x2 ANOVA results for negative moral agency, positive moral agency, and moral patiency for the AI or human assistant
}
\begin{tabular}{lllllll}
\hline
Dependent Variable    & Independent Variable & \textit{F} & \textit{df1} & \textit{df2} & \textit{p}      & \textit{$\eta^2_p$} \\
\hline
Negative moral agency & Valence              & 1254.44    & 1            & 680          & \textless .001* & .65                                 \\
                      & Agent type           & 37.47      & 1            & 680          & \textless .001* & .05                                 \\
                      & Valence x agent type & 0.31       & 1            & 680          & .58             & .00                                 \\
Positive moral agency & Valence              & 498.17     & 1            & 680          & \textless .001* & .42                                 \\
                      & Agent type           & 5.24       & 1            & 680          & .02*            & .00                                 \\
                      & Valence x agent type & 17.39      & 1            & 680          & \textless .001* & .02                                 \\
Moral patiency        & Valence              & 212.00     & 1            & 680          & \textless .001* & .24                                 \\
                      & Agent type           & 162.00     & 1            & 680          & \textless .001* & .19                                 \\
                      & Valence x agent type & 5.00       & 1            & 680          & .03*            & .01   
                      \\
\hline
\end{tabular}
\tablefootnote{* Significant \textit{p}-values (\textit{p} < .05)}
\label{tab:s2table1}
\end{table}

\paragraph{Agent’s positive moral agency}

There was a significant main effect of action valence on the agent’s perceived positive moral agency, supporting \textit{H1b}. There was also a significant effect of agent type and agent type-valence interaction. Post-hoc Tukey’s HSD tests showed that positive moral agency decreased in the immoral compared to the morally neutral condition for AI-assigned participants (immoral: $M$ = 3.16, SE = 0.10; morally neutral: $M$ = 4.71, SE = 0.09, $p$ < .001) and human-assigned participants (immoral: $M$ = 2.70, SE = 0.07; morally neutral: $M$ = 4.97, SE = 0.08, $p$ < .001), supporting \textit{H1b} for each group. There was also a significant difference between the group-specific effects of valence, such that positive moral agency decreased more for the human than the chatbot agent ($p$ < .001). (\Cref{tab:s2table1} and \Cref{fig:s2agentB})

\paragraph{Agent’s moral patiency}

There was a significant main effect of action valence on the agent’s moral patiency, supporting \textit{H1c}. There was also a significant effect of agent type and agent type-valence interaction. Post-hoc Tukey’s HSD tests showed that moral patiency decreased in the immoral compared to the morally neutral condition for AI-assigned participants (immoral: $M$ = 3.64, SE = 0.12; morally neutral: $M$ = 4.89, SE = 0.10, $p$ < .001) and human-assigned participants (immoral: $M$ = 4.76, SE = 0.09; morally neutral: $M$ = 6.45, SE = 0.05, $p$ < .001), supporting \textit{H1c} for each group. There was also a significant difference between the group-specific effects of valence, such that moral patiency decreased more for the human than the chatbot agent ($p$ < .001). (\Cref{tab:s2table1} and \Cref{fig:s2agentA})

\subsubsection{Spillover effects (\textit{H2}) and spillover asymmetry (\textit{H3})}

\begin{quote}
    \textit{A human or AI agent’s immoral action (a) increases attribution of negative moral agency, (b) decreases attribution of positive moral agency, and (c) decreases attribution of moral patiency to the group the agent belongs to (humans or AIs), relative to the agent’s morally neutral action (\textit{H2}). An AI agent's immoral action spills over to attributions of (a) negative moral agency, (b) positive moral agency, and (c) moral patiency to the AI group differently than a human agent's immoral action spills over to attributions to the human group (\textit{H3}).}
\end{quote}

\paragraph{Group’s negative moral agency}

\begin{figure}[ht]
    \includegraphics[width=\textwidth]{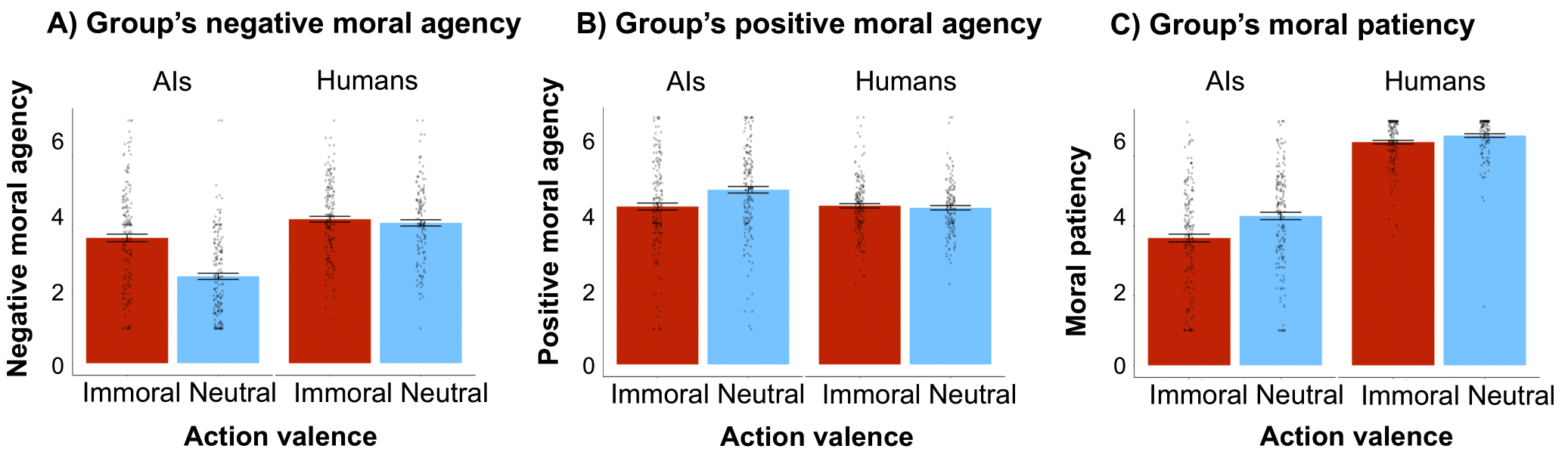}
    \caption{Effect of agent type and action valence on the moral attributions to AIs and humans in general. Bars represent the mean and error bars represent the standard error of the mean. Dots represent individual data points. A) Negative moral agency increased in the immoral condition compared to the morally neutral condition, more so for AIs than humans. B) Positive moral agency decreased in the immoral condition compared to the morally neutral condition, more so for AIs than humans. C) Moral patiency decreased in the immoral condition compared to the morally neutral condition, more so for AIs than humans.}
    \label{fig:s2group}
    \phantomsubcaption\label{fig:s2groupA}
    \phantomsubcaption\label{fig:s2groupB}
    \phantomsubcaption\label{fig:s2groupC}
\end{figure}

In the 2x2 model between agent type and action valence, there was a significant main effect of action valence on negative moral agency, supporting \textit{H2a}. Post-hoc Tukey’s HSD tests showed that negative moral agency increased in the immoral compared to the morally neutral condition for AIs in general (immoral: $M$ = 3.61, SE = 0.11; morally neutral: $M$ = 2.51, SE = 0.09, $p$ < .001), supporting \textit{H2a} for the AI group. However, contrary to \textit{H2a} for the human group, negative moral agency did not increase in the immoral compared to the morally neutral condition for humans in general (immoral: $M$ = 4.15, SE = 0.08, morally neutral: $M$ = 4.05, SE = 0.09, $p$ = .99). The agent type-valence interaction was also significant, supporting \textit{H3a}. The AI agent’s immoral action increased negative moral agency relative to the neutral action for AIs ($p$ < .001) and this was not observed for humans ($p$ = .99), meaning that there was an asymmetry in the spillover of negative moral agency. Lastly, there was also a significant effect of agent type. AIs had significantly lower overall negative moral agency than humans (AIs: $M$ = 3.05, SE = 0.08; humans: $M$ = 4.10, SE = 0.06). (\Cref{tab:s2table2} and \Cref{fig:s2groupA})

\begin{table}[ht]
\centering
\caption{Study 2 2x2 ANOVA results for negative moral agency, positive moral agency, and moral patiency for AIs or humans in general
}
\begin{tabular}{lllllll}
\hline
Dependent Variable    & Independent Variable & \textit{F} & \textit{df1} & \textit{df2} & \textit{p}      & \textit{$\eta^2_p$} \\
\hline
Negative moral agency & Valence              & 46.30      & 1            & 680          & \textless .001* & .06                                 \\
                      & Agent type           & 125.60     & 1            & 680          & .001*           & .16                                 \\
                      & Valence x agent type & 28.50      & 1            & 680          & \textless .001* & .04                                 \\
Positive moral agency & Valence              & 7.52       & 1            & 680          & \textless .001* & .01                                 \\
                      & Agent type           & 8.81       & 1            & 680          & \textless .001* & .01                                 \\
                      & Valence x agent type & 28.50      & 1            & 680          & \textless .001* & .04                                 \\
Moral patiency        & Valence              & 22.79      & 1            & 680          & \textless .001* & .03                                 \\
                      & Agent type           & 800.93     & 1            & 680          & \textless .001* & .54                                 \\
                      & Valence x agent type & 6.25       & 1            & 680          & .01*            & .00     
                      \\
\hline
\end{tabular}
\tablefootnote{* Significant \textit{p}-values (\textit{p} < .05)}
\label{tab:s2table2}
\end{table}

\paragraph{Group’s positive moral agency}

There was a significant main effect of action valence on positive moral agency, supporting \textit{H2b}. Post-hoc Tukey’s HSD tests showed that positive moral agency decreased in the immoral compared to the morally neutral condition for AIs in general (immoral: $M$ = 4.73, SE = 0.10; morally neutral: $M$ = 4.95, SE = 0.09, $p$ < .001), supporting \textit{H2b} for the AI group. However, contrary to \textit{H2b} for the human group, positive moral agency did not decrease in the immoral compared to the morally neutral condition for humans in general (immoral: $M$ = 4.50, SE = 0.06, morally neutral: $M$ = 4.44, SE = 0.06, $p$ = .87). The agent type-valence interaction was also significant, supporting \textit{H3b}. The AI agent’s immoral action decreased positive moral agency relative to the neutral action for AIs ($p$ < .001) and this was not observed for humans ($p$ = .87), meaning that there was an asymmetry in the spillover of positive moral agency between the groups. Lastly, there was also a significant effect of agent type. AIs had significantly higher overall positive moral agency than humans (AIs: $M$ = 4.71, SE = 0.07; humans: $M$ = 4.47, SE = 0.04). (\Cref{tab:s2table2} and \Cref{fig:s2groupB})

\paragraph{Group’s moral patiency}

There was a significant main effect of action valence on moral patiency, supporting \textit{H2c}. Post-hoc Tukey’s HSD tests showed that moral patiency decreased in the immoral compared to the morally neutral condition for AIs in general (immoral: $M$ = 3.65, SE = 0.11; morally neutral: $M$ = 4.28, SE = 0.10, $p$ < .001), supporting \textit{H2c} for the AI group. However, contrary to \textit{H2c} for the human group, moral patiency did not decrease in the immoral compared to the morally neutral condition for humans in general (immoral: $M$ = 6.34, SE = 0.05, morally neutral: $M$ = 6.59, SE = 0.05, $p$ = .47). The agent type-valence interaction was significant, supporting \textit{H3c}. The AI agent’s immoral action decreased moral patiency relative to the neutral action for AIs ($p$ < .001) and this was not observed for humans ($p$ = .47), meaning that there was an asymmetry in the spillover of moral patiency. Lastly, there was also a significant effect of agent type. AIs had significantly lower overall positive moral agency than humans (AIs: $M$ = 3.97, SE = 0.08; humans: $M$ = 6.49, SE = 0.04). (\Cref{tab:s2table2} and \Cref{fig:s2groupC})

\subsubsection{Discussion}

Consistent with Study 1 and \textit{H1}, in Study 2 we found that the immoral action of an AI or human agent increased the attribution of negative moral agency and decreased the attribution of positive moral agency and moral patiency to the agent relative to a neutral action. These attributions spilled over from the AI agent to AIs in general, as evidenced by the significant interactions, but this was not observed for attributions of the human agent to humans in general, partially supporting \textit{H2} (i.e., AI moral spillover but not human moral spillover). We also found an asymmetry between the spillover of moral attributions to AIs and humans, in which AIs were judged more harshly than humans after the moral transgression of one agent, which supports \textit{H3}. From an HCI perspective, this suggests that moral transgressions of one AI might not only taint the perception of similar AIs, but of AIs in general, with possibly detrimental consequences for cooperative HCI.

The fact that moral attributions of an AI might generalize to all AIs emphasizes the importance of careful design of AI agents to minimize the possibility that perceptions of AIs will be tainted because of one AI’s wrongdoings. This will require caution because we do not find comparable effects for human-human interaction, which has been the much more common context to date. As stated before, a possible explanation for this is that a minimal threshold of similarity must be reached for spillover to occur in humans \cite{uhlmann12}, suggesting the existence of boundary conditions for spillover to occur among humans. The superordinate human category might be perceived as less homogeneous than AIs, reducing spillover from one human agent to all humans. Interestingly, this also implies that AIs in general are perceived as a homogeneous group even when the AI agent is individuated.

\section{General discussion}

The present studies investigated moral spillover in human-AI interaction as compared to human-human interaction. We ran two experiments in which we assessed how moral attributions to a human or AI agent spill over to the group to which the agent belongs. In both experiments, we found that an agent’s immoral action increased attributions of negative moral agency and decreased attributions of positive moral agency and patiency to the agent compared to a morally neutral action (\textit{H1}). Moral attributions also spilled over from the AI agent to their group in both studies (chatbot assistants in Study 1, AIs in general in Study 2), and from the human agent to the human assistants group in Study 1 but not to the humans in general group in Study 2 (\textit{H2}). Therefore, the spillover asymmetry hypothesis (\textit{H3}) was only supported in Study 2, which had an individuated agent and a broader scope, engendering less similarity (i.e., more contrast) between the agent and the group. These are the first studies to show the effect of seemingly immoral behavior on attributions of moral patiency and to show moral spillover of agency or patiency in the AI context, which is a necessary step in understanding how increasingly prevalent and diverse AIs will fit into HCI. We summarize the implications for research and design in three contributions.

\subsection{AI moral spillover}

The finding of moral spillover builds on several streams of literature in human-AI interaction. For example, our findings are similar to \citet{do23a}, which found that people who interacted with a conversational assistant that made factual errors were less likely to continue participating in group discussions. In general, we show how the phenomenon of “algorithm aversion” \cite{dietvorst15a} is echoed by similar dynamics in the moral context, including a direct extension of \citet{longoni23}, which found a spillover of algorithm aversion in the context of government welfare programs. The spillover of moral patiency seems particularly understudied. Previous research has shown that people cared less about the wellbeing of immoral humans \cite{yu23}. Here, we showed that this also extends to AIs. There is a quickly growing literature on the moral patiency of AI that should incorporate the possibility of spillover from particular AIs to groups \cite{arrambide22a, harris21a, kahn12a, pauketat22b}.

\subsection{The AI double standard}

Our primary finding was the existence of moral spillover from the AI agent to the AI group, which occurred to both a specific AI group (chatbots) and to AIs in general. In this section, we speculate on why we found an "AI double standard" in Study 2 but not in Study 1, by which we mean a difference in moral spillover between the human and AI groups despite both agents taking negative actions. Specifically, in Study 2, we found that AIs were judged more harshly than humans for the actions of one agent. We tentatively speculate that this may be explained by differences in the similarity of an agent to their respective group across the two studies and between humans and AIs. Spillover was observed for three out of four conditions across the two studies (i.e., AIs in Study 1, humans in Study 1, and AIs in Study 2) but not in the fourth (i.e., humans in Study 2). There are two types of differences that could underlie this effect: first, differences between perceptions of AIs and humans that explain the human-AI difference, and second, methodological differences between the studies that explain the difference between Studies 1 and 2.

\subsubsection{Differences between perceptions of AIs and humans}

We draw on the psychological literature to propose an explanation of why people react different to humans and AIs. Spillover in humans seems to occur when the human group reaches a minimal threshold of similarity to the agent, such as kinship \cite{uhlmann12}. Indeed, in Study 1 we observed spillover for humans who had the same occupation as the human agent. In Study 2, the lack of spillover to all humans can be interpreted in light of psychological findings that people "represent humans as a heterogeneous group" \mbox{\cite{longoni23}}. Ingroup members are perceived as more diverse than outgroups \mbox{\cite{linville97, ostrom92, sedikides93}}, and therefore participants in Study 2 could have perceived all humans, a group they are also part of and that they are very familiar with (e.g., friends, family, public figures), as more diverse and less similar to the human agent. This reduced similarity between the human agent and human group could underpin the lack of spillover to all humans in Study 2.

On the other hand, people might readily and persistently categorize all AIs as one group of similar, homogeneous entities—and thus all are affected by one AI’s actions, explaining the AI spillover in both Studies 1 and 2. We speculate that humans currently lack sufficient experience with AIs to have stable, preconceived moral perceptions of AIs, as well as finding it difficult to distinguish between different AI kinds and competencies, unlike their stable expectations for humans as a whole and their tendency to perceive humans as a heterogeneous group with well-known variation, such as from the beloved moral leaders to the notorious villains of history. This is in line with \mbox{\cite{longoni23}}, where the generalization of AI errors was explained by the fact that "participants viewed algorithms as part of a more homogeneous group than people." Additionally, people might have clear expectations for how AIs should behave (e.g., as helpful and competent assistants), such that this schema collapses if one AI deviates. People also expect AIs to be more competent and to act less immorally than humans \cite{alon-barkat23, kocielnik19, rebitschek21, tolmeijer22a}, which could make the negative action more surprising. This might especially be the case in organizational contexts if AI systems are expected to perform to a minimum competency standard in order for the organization to opt into using them. In any case, the ease with which spillover occurs among AIs could have important implications for human-AI interaction. For instance, users could have reduced trust in AI that curtails beneficial interactions with trustworthy AI systems based on exposure to the mistakes or deliberate harm of one bad apple—as we discuss in more detail below.

\subsubsection{Methodological differences between Study 1 and Study 2}

The psychological literature may explain the different perceptions of AIs and humans, and the different perceptions of humans across Study 1 and Study 2 likely resulted from methodological differences. We posit three methodological differences that reduced the similarity between the human agent and group in Study 2 as compared to Study 1, thereby mitigating spillover:

\begin{enumerate}
    \item In Study 1, the groups were limited to the "assistant" profession (i.e., "AI-based chatbot assistant" or human "personal assistants"). In Study 2, the groups were broadened to "AIs, in general" and "humans, in general." This switch from an organizational context (all assistants) to a wider social context (all humans) might have helped prevent spillover for humans in Study 2 because it accentuated the difference between the human agent and the human group.
    \item In Study 2, individuating the agent with a name ("Ezal") may have further thus reduced the similarity between the human agent and the human group given the notion of the agent as a specific individual.
    \item Study 2 participants rated both human and AI groups, not just the group the agent belongs to as in Study 1. This simultaneous contrast between human and AI groups could have evoked feelings of human uniqueness, whereby each human seems more unique relative to each other, thus further reducing similarity between the human agent and the group \mbox{\cite{santoro23}}.
\end{enumerate}

These methodological differences also apply to the AI agent and group across Study 1 and Study 2, but they were not sufficient to curtail spillover. In the previous section, we speculated that this was because of the perception differences across AIs and humans: namely, the well-established psychological phenomenon of human ingroup heterogeneity in comparison to the apparent AI outgroup homogeneity \mbox{\cite{linville97, longoni23, ostrom92, sedikides93}}. Thus, our results can be understood as the methodological differences between Study 1 and Study 2 adding to the perception differences between humans and AIs, leading the only of the four conditions in which spillover did not occur to be the human-assigned participants of Study 2.

\subsection{Boundary conditions of blame and responsibility}

While our focus was the spillover effects to groups, we found that, in Study 1, attitudes towards the AI agent changed more than attitudes towards the human agent after their immoral behavior. This is in line with previous studies showing that AIs are judged more harshly than humans \cite{franklin21, hong20a, liu19a,liu21a}. However, the agent-specific findings of Study 2 contrasted with Study 1 and with prior literature. First, we found that the AI agent was attributed less negative moral agency and more positive moral agency, on average, than the human agents. Second, immoral behavior caused larger reductions in positive moral agency and moral patiency attributed to the AI agents than to the human agents.

These discrepancies could be partially explained by the fact that in Study 2 the AI agent was given a name, which could be quite unusual and have increased the personification of the agent and the extent to which people relate to the AI agent, while having a name is a feature of every human being. Increasing the humanization of AIs has been promoted in HCI as a way to increase positive attitudes towards AIs and the adoption of AI systems \cite{fink12, sebo20, seeger21}. Another possible explanation is that the AI agent in Study 1 was a chatbot, whereas the agent in Study 2 was not specified as a particular type of AI. It is possible that participants had greater familiarity with or specific biases towards chatbots, and thus judged them more harshly due to specific expectations of behavior. By contrast, they could have been more lenient towards the unspecified AI agent in Study 2 due to a lack of such expectations.

\section{Limitations and future research}

Our studies show the existence of moral spillover in human-AI interaction. However, we did not explicitly test the possible mechanisms through which moral spillover occurs. There are many factors that could be at play, such as familiarity with AI agents, similarity between an agent and the spillover target group, and pre-existing attitudes towards AI. Future research could investigate factors that affect moral spillover to gain a better understanding of how this phenomenon can take place in various HCI settings, such as by systematically varying scenarios to identify more specific boundary conditions. Further research could also investigate ways to prevent or mitigate spillover in order to avoid undue negative attributions towards groups of AIs.

In both studies, approximately 18\% of recruited participants in each study were excluded, following preregistered criteria, due to failing a reading comprehension check that asked them to identify the agent’s type. More participants failed this check than we expected and this may have skewed the sample towards particularly attentive participants, and it may indicate a challenge in ensuring participants grasp novel, hypothetical scenarios in which AIs are described like humans (e.g., “solves problems,” “does math calculations"). This confusion could be explained by scripts, schemas, or core beliefs about the roles and capabilities of AIs, which led to assumptions about the agent’s identity. Future research could delve into this with different vignettes or different comprehension checks, including asking participants to explain why they believe the agent is the type of agent they selected.

We tested moral spillover in a particular organizational setting. While we selected this setting because of its generic, widespread potential application (e.g., “department,” “work deadline,” “download specialized software”), there may still be idiosyncrasies that steered participant reactions. Additionally, unlike the vignettes in our studies, AI actions can be morally positive. Future studies could investigate how humans and AIs in different modern, realistic settings are perceived or how moral attributions spill over to different kinds of AIs that are currently widespread in society, including various service robots or conversational agents, such as ChatGPT. In addition to scenario variation, the effects we found could also be tested for persistence over time (e.g., in an actual follow-up test weeks or months after the initial intervention), in scenarios where the AI agent performs a moral action (e.g., helping their colleagues to improve work quality), or in real-world settings (e.g., in a controlled environment such as a university course where interaction can be generated inconspicuously). By understanding how spillover might affect a vast number of AIs with different applications, competencies, and levels of sophistication, we might be able to promote high-quality human-AI interactions, increase the acceptance of AIs in organizational or cooperative team settings and work towards harmonious relations between AIs and humans \cite{berberich20}. 

The existence of spillover for AIs emphasizes the need to design highly competent AIs while minimizing the possibility of error in human-AI interaction, since the negative actions of one AI could taint perceptions of all AIs. Additionally, designing collaborative AIs to be friendlier or warmer might balance out the perceptions of AI that produce harsh judgments of erring AIs. Designers could also make clear to humans working in cooperative human-AI settings the separability between different kinds of AI, which each have their own competencies and applications; this could reduce unwarranted spillover of perceptions of one AI to other kinds. AI design could also incorporate AI apologies for mistakes, especially those perceived as intentional or immoral, to increase trust in AI team members \mbox{\cite{buchholz17, lopez23, zhang23}}. Finally, designs could be transparent about AI imperfections and the potential need for collaborative human-AI teamwork to reduce harsher judgments of AIs and repair broken trust \mbox{\cite{schelble22, shank19, zhang24}}.

\section{Conclusion}

The present study demonstrated the spillover of moral agency and patiency in an HCI setting. We found spillover from one chatbot assistant to all chatbot assistants and from an individuated AI agent to all AIs. However, moral spillover to AI groups may have important differences to the well-documented phenomenon of moral spillover to human groups. In Study 2, we found evidence of a moral double standard, in which moral attributions to AIs in general are affected more by an AI agent’s immoral action than moral attributions to humans in general are affected by a human agent’s action. Moral spillover may be an important dynamic as a variety of AI systems become more common in everyday life, particularly in contexts with significant moral consequences, and more research is needed to ensure the design of AI systems that support accurate human perceptions and reasonable judgments of AI behavior.


\bibliographystyle{ACM-Reference-Format}
\bibliography{bibliography/spillover}

\end{document}